\documentclass[11pt]{article}

\usepackage{acl}
\usepackage{placeins}
\usepackage{float}
\usepackage{times}
\usepackage{latexsym}
\usepackage[T1]{fontenc}
\usepackage[utf8]{inputenc}
\usepackage{microtype}
\usepackage{inconsolata}
\usepackage{graphicx}
\usepackage{stfloats}
\usepackage{booktabs}
\usepackage{multirow}
\usepackage{amsmath}
\usepackage{url}
\usepackage{siunitx}
\usepackage{tcolorbox}
\tcbuselibrary{skins,breakable}

\title{When Built-in Thinking Helps and Hurts: Constraint-Level Error Shifts in Instruction Following}

\author{
Sai Adith Senthil Kumar \\
George Mason University \\
\texttt{ssenthi2@gmu.edu}
}

\begin{document}
\maketitle

\begin{abstract}
Large reasoning models (LRMs) often improve math and coding performance, but their effect on \textbf{instruction following} is unclear.
We study IFEval with Qwen3 models (1.7B--32B), using same-weights Thinking ON/OFF controls; four Hunyuan models provide directional cross-family support.
Aggregate pass-rate changes are small ($-0.55$ to $-3.52$~pp), yet 10--20\% of prompts switch between pass and fail across modes, suggesting that thinking changes the pattern of errors---some prompts improve while others worsen---rather than uniformly degrading performance.
Under a post-hoc Qwen3-derived grouping, constraint types separate into \textbf{Planning} (global counting, structure, coordination), which improves at the class level under thinking, and \textbf{Precision} (exact local form), which consistently worsens; the class-level Planning/Precision sign pattern holds directionally for all four Hunyuan models despite Hunyuan's opposite aggregate direction.
Thinking also changes final-answer length; matched-length analyses substantially reduce the Precision drop, but a residual penalty remains.
Analyzing thinking traces with a cross-encoder relevance metric reveals three patterns: \textbf{Neutral} shows a positive relevance--compliance link ($r \approx 0.15$); \textbf{Planning} shows near-zero predictive correlation ($r \approx 0.02$) despite measurable trace engagement, consistent with an \textbf{execution gap} between CE-measured trace relevance and final-answer compliance; \textbf{Precision} shows a small negative correlation ($r \approx -0.05$), with failing instances having higher mean relevance than passing ones.
Activation patching across four model sizes (1.7B--14B) shows that Precision flip instances are more often restored than Planning flip instances (32--58\% vs.\ 14--40\% mean layer-restoration), with the largest gap at 14B (about 30\,pp).
\end{abstract}

\section{Introduction}

Chain-of-thought (CoT) reasoning helps on math, code, and symbolic tasks \citep{wei2022cot,nye2021scratchpad,sprague2024tocot}. Instruction-following benchmarks such as IFEval, FollowBench, and InFoBench \citep{zhou2023ifeval,jiang2024followbench,qin2024infobench} test a different ability: whether models can satisfy natural-language instructions and constraints. We focus on IFEval, whose constraints are explicit and checkable, e.g., ``no commas,'' ``write exactly 100 words,'' or ``always use uppercase.'' These constraints are often easy to understand but unforgiving; success depends less on deduction than on precise final-answer control.

Recent work shows that reasoning can degrade instruction adherence in both general and math-specific instruction-following settings \citep{li2025whenthinkingfails,fu2025scaling}. We focus on whether built-in thinking changes errors at the constraint-type level and whether such patterns hold across model families. We ask: \textbf{does built-in thinking improve instruction following, and how does it change error types?} We use Qwen3 models \citep{yang2025qwen3} as the primary controlled setting: Thinking ON/OFF uses identical weights and prompts, so the inference-time thinking-mode signal is the primary controlled difference. We then test whether the resulting constraint-level pattern holds directionally in independent Hunyuan models.\footnote{\url{https://huggingface.co/collections/tencent/hunyuan-dense-model} and \url{https://huggingface.co/tencent/Hunyuan-A13B-Instruct}.}

Our contributions are: (i) Thinking ON changes the pattern of IFEval errors rather than uniformly helping or hurting: 10--20\% of prompts flip pass/fail, with \textbf{Planning} constraints improving and \textbf{Precision} constraints worsening; the Qwen3-derived Planning/Precision directional split holds for all four Hunyuan models despite Hunyuan's opposite aggregate sign; (ii) matched-length filtering reduces the Precision penalty from $-8.5$~pp to between $-1.6$ and $-2.3$~pp, so output length accounts for much, but not all, of the drop in this robustness analysis; (iii) cross-encoder trace relevance is consistent with an \emph{execution gap}: Planning traces contain CE-measured rule-relevant content, but this relevance does not predict compliance ($r \approx 0.02$); and (iv) activation patching shows Precision flip instances are more often restored than Planning flip instances across four Qwen3 sizes, with the largest gap at 14B (about 30\,pp).

\section{Related Work}

\paragraph{Instruction following and CoT.}
IFEval \citep{zhou2023ifeval} evaluates instruction following with deterministic, verifiable constraints; related benchmarks study fine-grained constraint following \citep{jiang2024followbench}, broader instruction-following ability \citep{qin2024infobench}, and multi-constraint composition \citep{wen2024complexbench}. Together, these benchmarks support the view that instruction following is not a single scalar ability but a mixture of constraint types and composition demands. Other work examines counterfactual instruction following, showing that models can fail to comply with instructions that require intentional underperformance, highlighting failures rooted in trained behavioral priors rather than constraint misunderstanding \citep{kumar2025counterfactual}. Our work studies a different setting: same-weights thinking-mode interventions on verifiable IFEval constraints, with constraint-level error redistribution, trace-relevance diagnostics, and activation-patching recoverability analyses. CoT prompting helps on multi-step reasoning \citep{wei2022cot,nye2021scratchpad} but gains are highly task-dependent \citep{sprague2024tocot}. IFEval's per-constraint pass/fail scoring makes it well suited for asking \emph{which} constraints shift under an intervention.

\paragraph{Thinking and instruction following.}
Prior work has explored both the benefits and risks of adding explicit reasoning to instruction-following models. Training models to generate thoughts before answering has been shown to improve general instruction following \citep{wu2025thinkingllms}, while reasoning can also degrade instruction adherence in general instruction-following settings \citep{li2025whenthinkingfails,qin2025raif}; related work studies this tradeoff in math-domain instruction following via MathIF \citep{fu2025scaling}. Other work examines whether reasoning traces themselves follow user instructions \citep{kwon2025reasonif}. Unlike CoT-prompting studies, our setup changes only the built-in thinking mode while keeping model weights and task prompts fixed. Our work differs on three axes: we use same-weights Thinking ON/OFF controls, derive a constraint-level error pattern on Qwen3, and test whether the Planning/Precision directional contrast holds in a second model family. Our trace-relevance analysis asks whether the trace \emph{engages} with the rule and whether engagement predicts final-answer compliance, rather than whether the trace itself satisfies the rule. This separates trace-level instruction adherence from the downstream question of whether reasoning helps the final answer obey the constraint. Together, these threads motivate asking not just whether thinking hurts or helps overall, but which constraint types drive each direction.

\section{Experimental Setup}
\label{sec:setup}

\subsection{Benchmark}
IFEval \citep{zhou2023ifeval} contains 541 prompts paired with constraints from 25 instruction types (counting, formatting, keyword, structural). Each constraint is scored deterministically as pass/fail. We report \textbf{prompt-level strict} accuracy (all constraints satisfied) and \textbf{instruction-level strict} accuracy (per-constraint pass rate).
All pass-rate deltas are reported in percentage points (pp).

\subsection{Models and modes}
We evaluate Qwen3-1.7B, 4B, 8B, 14B, and 32B as primary analysis models. Each is run in two modes: in \textbf{Thinking ON}, the model writes an internal reasoning trace inside \texttt{<think>...</think>} before the final answer; in \textbf{Thinking OFF}, it answers directly \citep{yang2025qwen3}. Both modes share the same weights and receive the same unmodified task prompts; only the inference-time chat-template signal differs, reducing the prompt-wording confound present in CoT-prompted comparisons.

We use greedy decoding throughout, with seed 22 as the main run. As a seed check, we also ran the Qwen3 ON/OFF comparison at seeds 11 and 33; every aggregate ON/OFF direction is preserved, with ON--OFF delta ranges of at most 0.55~pp prompt-level and 0.36~pp instruction-level. For scoring, we remove any generated \texttt{<think>...</think>} block and pass only the final answer to the official IFEval checker.

For directional cross-family support, we evaluate Hunyuan-1.8B, 4B, 7B, and A13B by toggling each public Hugging Face checkpoint between Thinking ON/OFF modes.\footnote{Dense checkpoints: \url{https://huggingface.co/collections/tencent/hunyuan-dense-model}; A13B checkpoint: \url{https://huggingface.co/tencent/Hunyuan-A13B-Instruct}.} Hunyuan is used only for directional cross-family analysis; all length, trace-relevance, and patching analyses use Qwen3.

\section{Results}

\begin{table*}[t]
\centering
\small
\setlength{\tabcolsep}{7pt}
\caption{IFEval prompt- and instruction-level strict accuracy with and without thinking
for both model families under greedy decoding.
$\Delta = \text{ON} - \text{OFF}$; negative $\Delta$ means thinking hurts.
Final answers are scored after removing generated thinking blocks.
Hunyuan results are used for directional cross-family support only.}
\label{tab:main_results}
\begin{tabular}{lrrrrrr}
\toprule
& \multicolumn{3}{c}{Prompt-level strict (\%)} & \multicolumn{3}{c}{Instruction-level strict (\%)} \\
\cmidrule(lr){2-4}\cmidrule(lr){5-7}
Model & Thinking OFF & Thinking ON & $\Delta$ & Thinking OFF & Thinking ON & $\Delta$ \\
\midrule
\multicolumn{7}{l}{\textit{Qwen3 (primary analysis)}} \\
\midrule
Qwen3-1.7B   & 62.48 & 58.96 & $-3.52$ & 70.26 & 67.27 & $-2.99$ \\
Qwen3-4B     & 73.01 & 70.61 & $-2.40$ & 79.62 & 77.10 & $-2.52$ \\
Qwen3-8B     & 75.05 & 73.75 & $-1.30$ & 80.94 & 79.38 & $-1.56$ \\
Qwen3-14B    & 76.52 & 75.97 & $-0.55$ & 82.73 & 82.13 & $-0.60$ \\
Qwen3-32B    & 76.16 & 73.94 & $-2.22$ & 82.01 & 80.22 & $-1.79$ \\
\midrule
\multicolumn{7}{l}{\textit{Hunyuan (directional cross-family support)}} \\
\midrule
Hunyuan-1.8B & 48.06 & 49.72 & $+1.66$ & 58.63 & 59.83 & $+1.20$ \\
Hunyuan-4B   & 61.37 & 64.70 & $+3.33$ & 71.46 & 72.66 & $+1.20$ \\
Hunyuan-7B   & 63.40 & 63.40 & $+0.00$ & 71.10 & 70.98 & $-0.12$ \\
Hunyuan-A13B & 70.98 & 72.46 & $+1.48$ & 77.46 & 79.26 & $+1.80$ \\
\bottomrule
\end{tabular}
\end{table*}

\subsection{Average changes are small, but many prompts flip}
\label{sec:aggregate}

Table~\ref{tab:main_results} shows prompt-level and instruction-level accuracy. Prompt-level Thinking ON reduces Qwen3 accuracy by only $0.55$ to $3.52$~pp, suggesting a modest aggregate effect. However, this average masks substantial disagreement: the \textbf{disagreement rate}, or fraction of prompts that flip pass/fail between Thinking ON and OFF, is 10--20\% for Qwen3 (Table~\ref{tab:disagree})---and the flips are not balanced.

\begin{table}[t]
\centering
\small
\caption{Prompt-level ON/OFF disagreement under IFEval scoring. Helped = OFF fails, ON passes; Hurt = OFF passes, ON fails.}
\label{tab:disagree}
\begin{tabular}{lccc}
\toprule
Model & \%~Flip & \%~Helped & \%~Hurt \\
\midrule
\multicolumn{4}{l}{\textit{Qwen3}} \\
\midrule
Qwen3-1.7B   & 20.1 & 8.3  & 11.8 \\
Qwen3-4B     & 13.5 & 5.5  &  7.9 \\
Qwen3-8B     & 11.6 & 5.2  &  6.5 \\
Qwen3-14B    & 10.2 & 4.8  &  5.4 \\
Qwen3-32B    & 11.5 & 4.6  &  6.8 \\
\midrule
\multicolumn{4}{l}{\textit{Hunyuan}} \\
\midrule
Hunyuan-1.8B & 24.4 & 13.1 & 11.3 \\
Hunyuan-4B   & 21.6 & 12.6 &  9.1 \\
Hunyuan-7B   & 23.5 & 11.8 & 11.6 \\
Hunyuan-A13B & 18.9 & 10.4 &  8.5 \\
\bottomrule
\end{tabular}
\end{table}

Within Qwen3, smaller models tend to flip more often and are also hurt more in aggregate, with 14B showing the smallest prompt-level drop ($-0.6$~pp).
The 32B result breaks a simple monotonic scale pattern, so scale is not the sole driver of the aggregate effect.

\paragraph{Cross-family comparison (Hunyuan).}
Hunyuan differs in aggregate: three Hunyuan models improve prompt-strict accuracy ($+1.48$ to $+3.33$~pp) and Hunyuan-7B is net-neutral ($+0.00$~pp), unlike every Qwen3 size.
Disagreement rates remain comparable (18.9--24.4\%), so substantial ON/OFF disagreement appears in both families even though the aggregate sign differs.

\subsection{Thinking changes errors across constraint types}
\label{sec:error-redistribution}

To characterize prompt flips, we examine per-constraint instruction-level deltas across the five Qwen3 sizes. We use a post-hoc descriptive grouping rather than a predefined taxonomy. We first identify constraint types with stable negative ON--OFF deltas as \textbf{Precision}. Among the remaining constraints, we group instruction types with positive or near-zero-positive Qwen3 deltas and clear global counting, structure, or coordination requirements as \textbf{Planning}; the rest are \textbf{Neutral}. This grouping summarizes the Qwen3 pattern plus semantic inspection, not a unique taxonomy; Hunyuan tests whether the resulting Planning/Precision directional contrast holds (§\ref{sec:hunyuan-validation}). Appendix Table~\ref{tab:classification} lists every instruction type and rationale.

\paragraph{Planning constraints (5 instruction types).}
These instruction types have positive or non-negative Qwen3 deltas in most sizes and share a planning-like structure: they require counting, coordinating parts of the response, or tracking a global property across the final answer. This suggests the reasoning trace provides a workspace, so we label this group \textbf{Planning}.

\paragraph{Precision constraints (9 instruction types).}
Instruction types where Thinking ON consistently hurts pass rate show a precision-like pattern: success depends on exact local form, and one stray token, punctuation mark, or boundary error can fail the constraint. Thinking appears less helpful for this kind of local control and may make it harder by increasing output length, so we label this group \textbf{Precision}.

\paragraph{Neutral (11 instruction types).}
The remaining constraints show no stable ON/OFF direction across model sizes; we label them \textbf{Neutral}.

These labels summarize the Qwen3 pattern: Thinking ON yields positive Planning class-level deltas, negative Precision deltas, and mixed Neutral effects. Appendix~\ref{app:qwen_breakdown} shows the full per-constraint breakdown. We next test whether this Planning/Precision contrast holds in Hunyuan.

Class-level results average over instruction types within each class rather than over raw constraint instances. This prevents frequent instruction types from dominating the class summary and keeps the unit of analysis aligned with the grouping procedure. Because the labels are assigned before looking at Hunyuan, the Hunyuan comparison tests transfer of the Qwen3-derived directional contrast rather than re-fitting a taxonomy to a second family.

The labels should therefore be read as descriptive bins for the observed intervention effect, not as claims that IFEval has only three underlying cognitive categories. The main claim is the directional split induced by thinking mode.

\subsection{Cross-family directional support: Hunyuan}
\label{sec:hunyuan-validation}

\begin{figure}[t]
\centering
\includegraphics[width=\columnwidth]{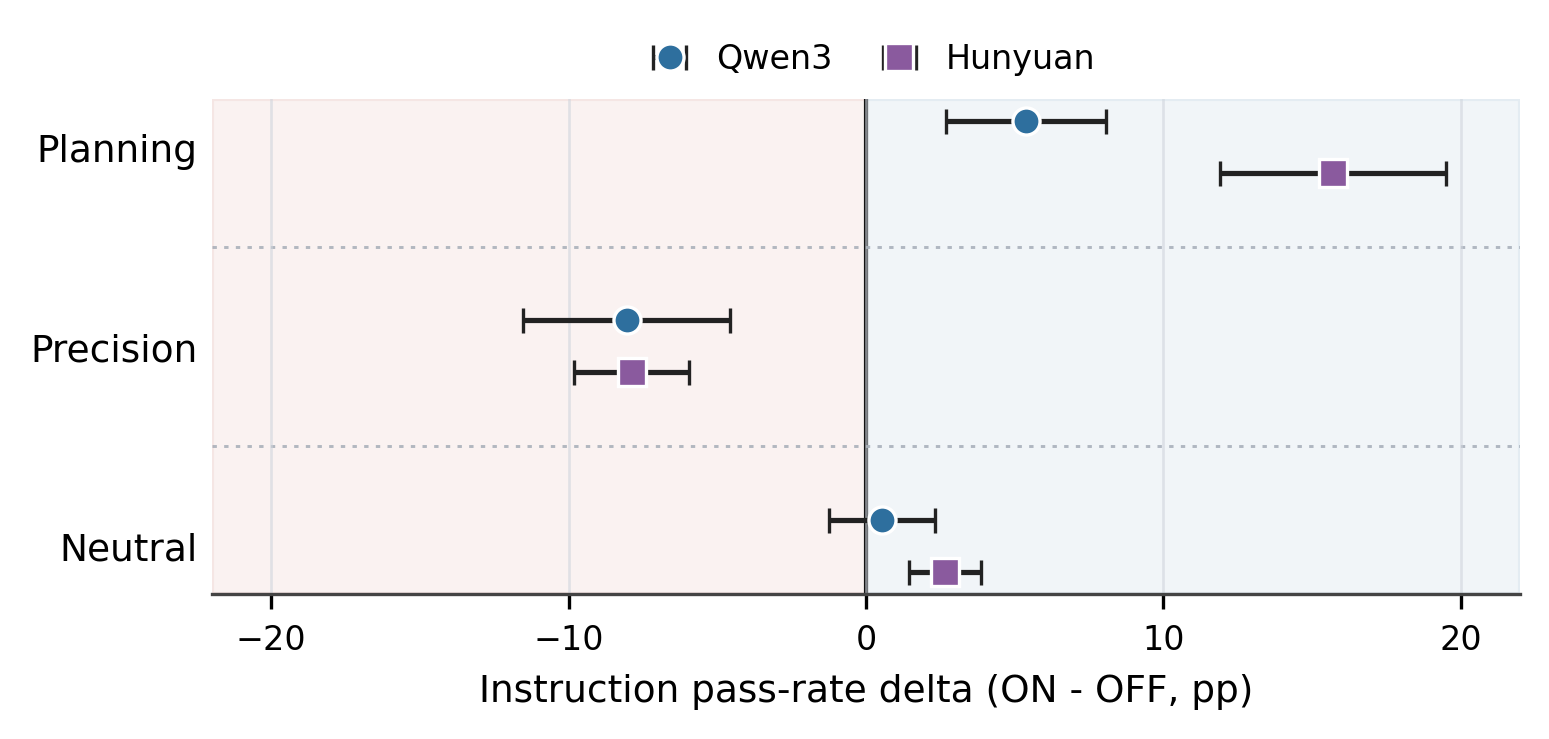}
\caption{Class-level instruction-type ON$-$OFF pass-rate deltas for Qwen3 and Hunyuan. Each model first contributes a mean over instruction types within each fixed class; markers show family means and error bars show 95\% CIs over models. Positive means Thinking ON helps.}
\label{fig:family_class_deltas}
\end{figure}

To test whether the Qwen3-derived class pattern holds directionally beyond Qwen3, we keep the labels fixed and apply them to the four Hunyuan models. Figure~\ref{fig:family_class_deltas} compares class-level deltas for Qwen3 and Hunyuan; Table~\ref{tab:hunyuan_class} reports the Hunyuan per-model values. Appendix~\ref{app:qwen_breakdown} shows the full Qwen3 and Hunyuan per-constraint breakdowns.

\begin{table}[t]
\centering
\small
\caption{Hunyuan instruction-level ON$-$OFF delta (pp) by fixed Qwen3-derived class.}
\label{tab:hunyuan_class}
\begin{tabular}{lrrr}
\toprule
Model & Planning $\Delta$ & Precision $\Delta$ & Neutral $\Delta$ \\
\midrule
Hunyuan-1.8B & $+20.1$ & $-9.6$  & $+4.1$ \\
Hunyuan-4B   & $+16.0$ & $-5.3$  & $+1.6$ \\
Hunyuan-7B   & $+10.6$ & $-9.2$  & $+1.6$ \\
Hunyuan-A13B & $+16.1$ & $-7.4$  & $+3.3$ \\
\bottomrule
\end{tabular}
\end{table}

All four Hunyuan models show Planning~$\Delta > 0$ and Precision~$\Delta < 0$, matching Qwen3 at the class-sign level across dense and MoE architectures. The largest swings occur for Hunyuan-1.8B ($+20.1$~pp Planning, $-9.6$~pp Precision). Hunyuan's Neutral class is positive, whereas Qwen3's is smaller and mixed, so cross-family support is strongest for the Planning/Precision directional contrast, not for magnitudes or every individual instruction type.

\subsection{Output length partially explains the Precision penalty}
\label{sec:length-control}

A natural question is whether the Precision drop is a length artifact: longer ON answers create more opportunities for small token-level mistakes.

\begin{table}[t]
\centering
\small
\setlength{\tabcolsep}{4pt}
\caption{Qwen3 final-answer word counts after removing thinking blocks.
Shifts are ON $-$ OFF.}
\label{tab:length_shift}
\begin{tabular}{lrrrr}
\toprule
Model & Med.\ OFF & Med.\ ON & Med.\ shift & Mean shift \\
\midrule
1.7B & 153 & 169 & $+16$ & $+67$ \\
4B   & 172 & 163 & $-9$  & $+34$ \\
8B   & 173 & 171 & $-2$  & $+14$ \\
14B  & 173 & 164 & $-9$  & $+11$ \\
32B  & 179 & 172 & $-7$  & $+21$ \\
\bottomrule
\end{tabular}
\end{table}

\paragraph{How much do final-answer lengths shift?}

Table~\ref{tab:length_shift} reports both median and mean shifts because they capture different effects: the median reflects the typical prompt-level change, while the mean captures occasional long ON answers. Only 1.7B has a clearly positive median shift ($+16$ words); larger models are near-zero or slightly negative, while mean shifts are positive for every model ($+11$ to $+67$ words). This is consistent with a length contribution, especially at 1.7B, but does not fully explain the pattern.

\paragraph{The class-specific pattern holds across length bins.}

\begin{figure}[t]
\centering
\includegraphics[width=\columnwidth]{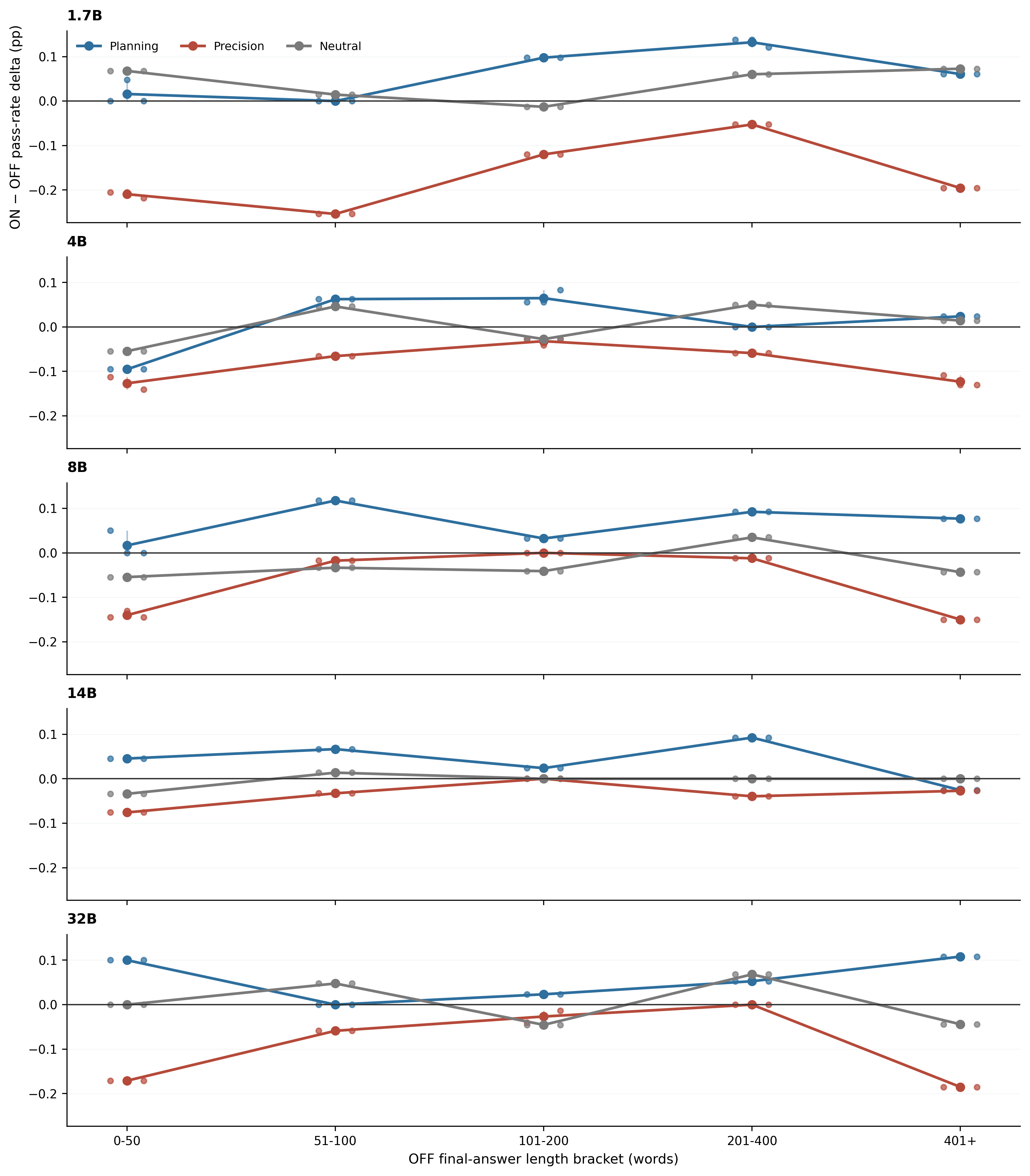}
\caption{Instruction-level ON$-$OFF pass-rate delta after grouping examples by OFF
final-answer length. OFF length is used as the baseline length before the thinking-mode
intervention; positive values mean Thinking ON helps.}
\label{fig:thinking_len}
\end{figure}

\begin{figure*}[!t]
\centering
\includegraphics[width=\textwidth]{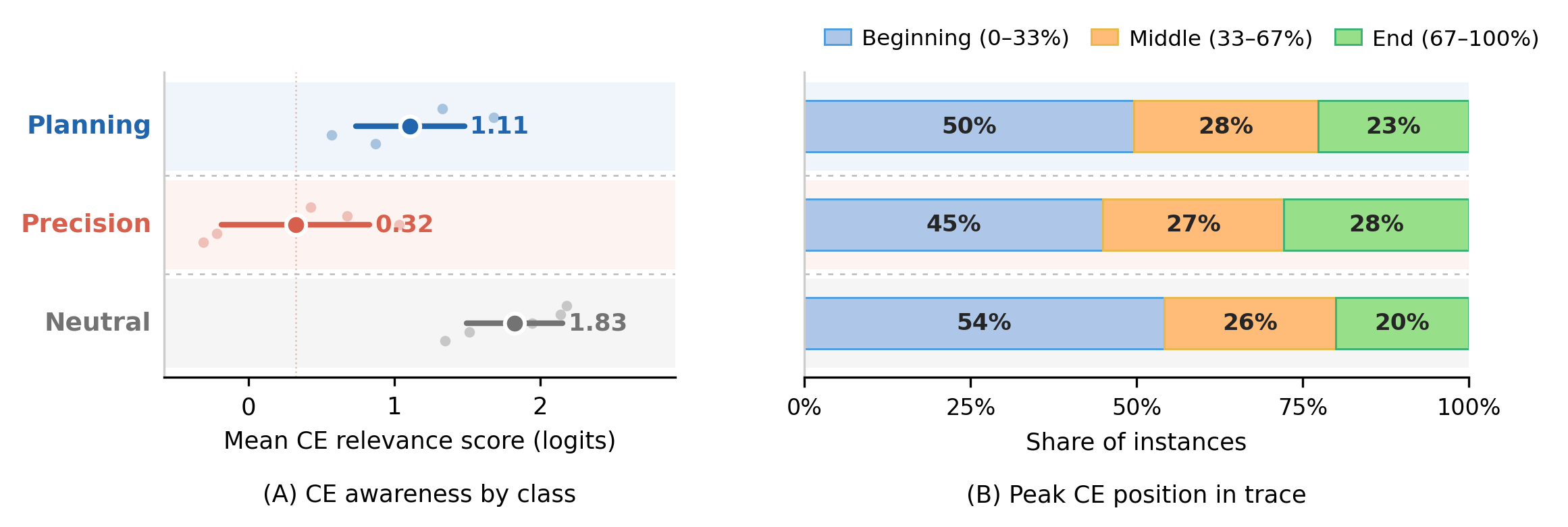}
\caption{\textbf{(A)} Mean Cross-Encoder (CE) relevance per constraint class, averaged across five Qwen3 sizes (dots = per-model means; bar = 95\% CI). Neutral is highest; Precision lowest. \textbf{(B)} Peak CE relevance position in the thinking trace. Across classes, peaks occur mostly in the first third, indicating early constraint engagement.}
\label{fig:ce_awareness_table}
\end{figure*}

As a diagnostic stratification, we group examples by OFF final-answer length, which is measured before the thinking-mode intervention changes the final answer. Planning deltas remain mostly positive and Precision deltas remain mostly negative across bins (Fig.~\ref{fig:thinking_len}), so the Planning/Precision split is not confined to a single baseline length range.

\paragraph{Matched-length robustness check.}

As a stronger descriptive control, we compare ON/OFF pairs with similar final-answer lengths. We restrict to ON/OFF instruction pairs whose final-answer word counts fall within a given relative threshold and recompute class deltas (Table~\ref{tab:matched_length}).

\begin{table}[t]
\centering
\small
\caption{ON$-$OFF pass-rate delta (pp) for all data and length-matched subsets, by
constraint class, pooled over Qwen3 instruction pairs.
Ret.\% = fraction of instruction pairs retained.}
\label{tab:matched_length}
\begin{tabular}{lrrrr}
\toprule
Threshold & Ret.\% & Planning & Precision & Neutral \\
\midrule
All data  & 100   & $+5.3$ & $-8.5$ & $+0.6$ \\
$\pm$30\% &  58   & $+4.5$ & $-1.6$ & $+0.0$ \\
$\pm$20\% &  43   & $+3.6$ & $-1.8$ & $+0.5$ \\
$\pm$10\% &  24   & $+4.7$ & $-2.3$ & $+1.9$ \\
\bottomrule
\end{tabular}
\end{table}

Across all three thresholds, Planning stays positive and Precision stays negative. The Precision penalty shrinks under matching ($-8.5$~pp unmatched, versus between $-1.6$ and $-2.3$~pp matched), so length accounts for a substantial share of the effect but is not complete. Because the retained subset excludes pairs where thinking most changes output length, these matched estimates are a robustness check rather than a causal estimate. At $\pm$20\%, retained shares are similar across classes; model-level retention rates are in Appendix Table~\ref{tab:matched_by_model}.

\paragraph{Thinking-trace length as a secondary descriptive check.}

Trace length is not a reliable proxy for ON-mode success: pooled correlations are weakly negative for Planning ($r \approx -0.10$), positive for Precision ($r \approx +0.18$), and near zero for Neutral ($r \approx 0.02$; Appendix Table~\ref{tab:trace_length_corr}). These weak, mixed directions suggest trace length alone does not explain the Planning/Precision split.

\subsection{Constraint relevance in the thinking trace}
\label{sec:awareness}

We ask how strongly the thinking trace aligns with the active constraint and whether alignment predicts compliance. We use \textbf{cross-encoder relevance} (CE) as the primary metric and cosine relevance as a secondary baseline.

\paragraph{Measuring trace relevance.}
For each instruction, we split the trace into overlapping windows and score each window against the constraint description; trace relevance is the maximum window score. This max-window measure indicates trace engagement, not average trace quality. CE relevance uses the \texttt{cross-encoder/ms-marco-MiniLM-L6-v2} SentenceTransformers checkpoint,\footnote{\url{https://huggingface.co/cross-encoder/ms-marco-MiniLM-L6-v2}.} trained on MS MARCO passage ranking \citep{bajaj2016msmarco}; because it is trained for search relevance rather than reasoning quality, we treat it as a proxy for constraint-relevant trace content. For comparison, cosine relevance uses SentenceTransformers embeddings \citep{reimers2019sentencebert} as a looser baseline (Appendix~\ref{app:cosine_awareness}).

\paragraph{How much does the trace engage, and where?}

Figure~\ref{fig:ce_awareness_table} summarizes both mean CE trace relevance and where peak relevance occurs. CE scores are uncalibrated logits, so we interpret them only comparatively: higher scores mean stronger constraint--trace relevance under the cross-encoder, not greater reasoning depth. Neutral has the highest CE relevance ($\approx 1.8$), Planning is intermediate ($\approx 1.1$), and Precision is lowest ($\approx 0.3$); true constraint--trace pairs score above shuffled pairs in 49/50 cases (Appendix Table~\ref{tab:sanity}). Peak CE relevance falls most often in the \emph{first third} of the trace for all classes (roughly $45$--$54\%$ of examples). Cosine yields the same class ordering but different peak positions; lexical-overlap checks are in Appendix~\ref{app:lexical_overlap}.

\paragraph{Does engagement predict success?}

\begin{figure}[t]
\centering
\includegraphics[width=\columnwidth]{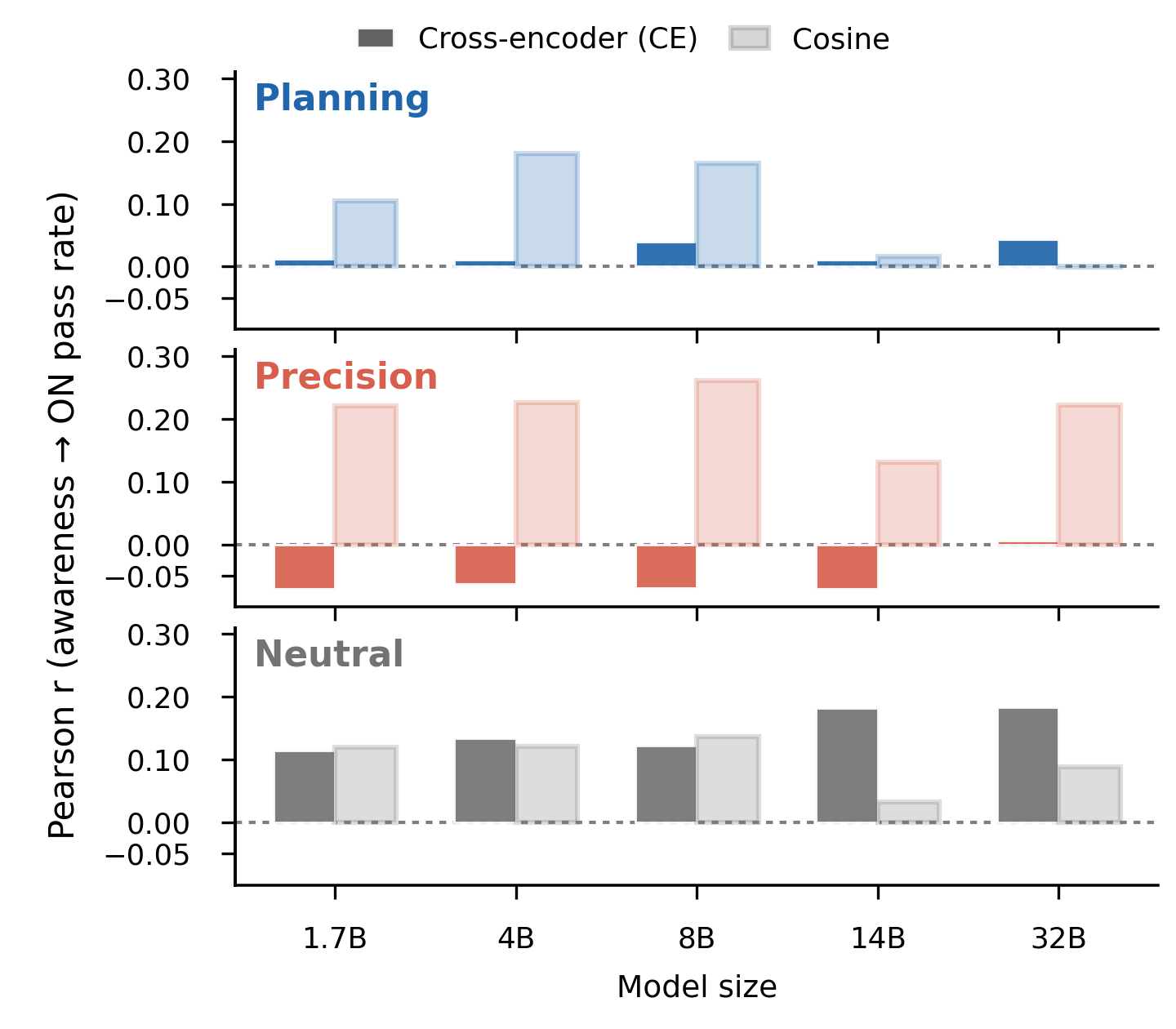}
\caption{Pearson $r$ between trace relevance and ON-mode pass, by class across Qwen3 sizes (1.7B--32B). Solid bars show CE (primary); transparent bars show cosine (secondary).}
\label{fig:ce_correlation}
\end{figure}

Figure~\ref{fig:ce_correlation} compares trace relevance with the ON-mode pass indicator. Under CE, \textbf{Neutral} is the clearest positive case: $r \in [0.10, 0.18]$ across sizes (mean $r = 0.148$). \textbf{Planning} remains near zero (mean $r \approx 0.02$), and \textbf{Precision} is weakly negative on average (mean $r \approx -0.05$). Precision is also the clearest metric-sensitive case: cosine suggests a positive relevance--success relation ($r \approx 0.21$), but CE does not (Appendix Table~\ref{tab:precision_metric_disagreement}). Thus, trace relevance is not uniformly predictive of final compliance; its relationship to success depends on the constraint class and metric.

\paragraph{Planning: the execution gap.} Thinking ON improves the Planning class overall (§\ref{sec:error-redistribution}), yet CE relevance is nearly uncorrelated with which instances succeed ($r \approx 0.02$). We treat this as evidence for an execution-gap hypothesis: traces often contain CE-measured rule-relevant content, but that content does not reliably carry through. This does not rule out qualitatively poor engagement, but it is inconsistent with a simple failure of awareness: the ON fail / OFF pass group has the highest mean Planning CE relevance ($2.82$; Appendix Table~\ref{tab:outcome_awareness}). Appendix~\ref{app:execution_gap} gives two cases.

\paragraph{Precision: weak negative engagement signal.}
Precision shows a related but different failure mode: the issue is local final-answer control rather than carrying out a global plan. Precision has the lowest mean CE engagement ($\approx 0.32$), and CE relevance is weakly negative on average, consistent with the length analysis (§\ref{sec:length-control}): for exact-form constraints, attending to the rule is not enough if the final answer later slips on a local token-level requirement.

\begin{figure*}[!t]
\centering
\includegraphics[width=0.9\textwidth]{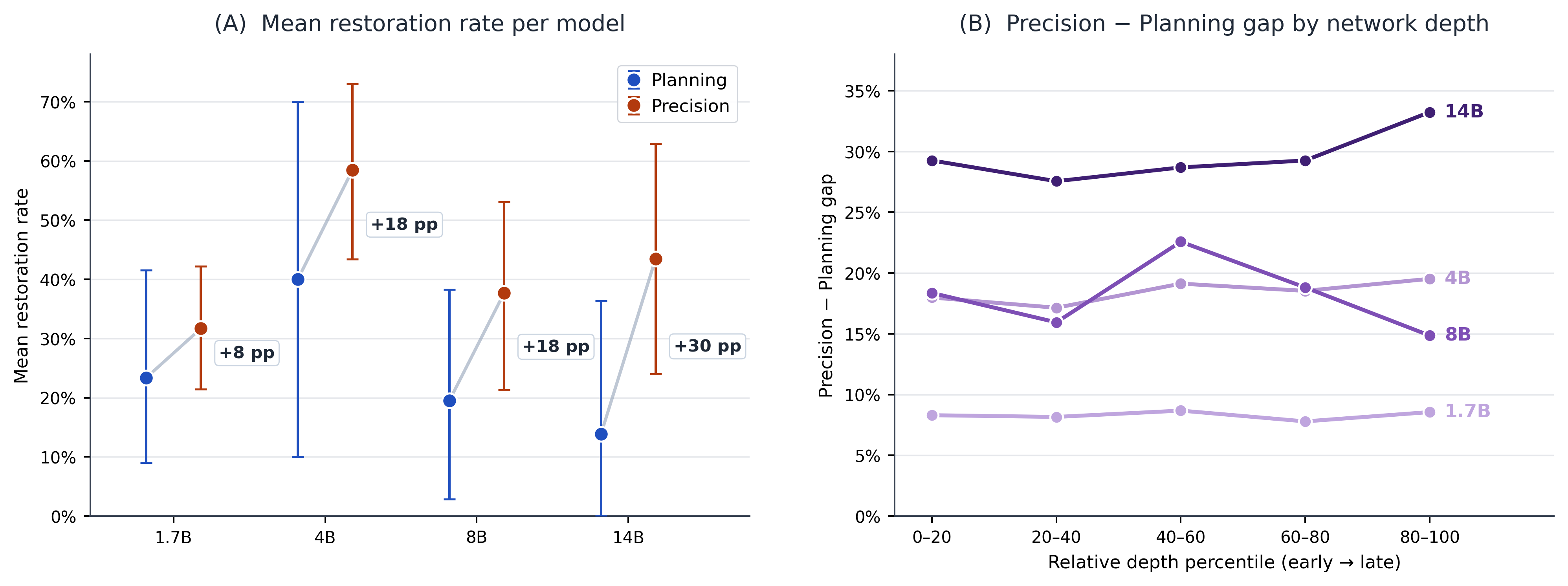}
\caption{%
  \textbf{Prefill activation patching restores Precision flip instances more often than Planning flip instances.}
  (a) Mean restoration rate per Qwen3 size, averaged over all layers and over attention and MLP components; error bars are 95\% prompt-bootstrap CIs over flip instances.
  (b) Precision$-$Planning restoration gap across five equal-width relative-depth bins, one line per model size.
  The gap is positive at every patched scale, is smallest at 1.7B and largest at 14B, and varies less across depth than across model size.%
}
\label{fig:patching}
\end{figure*}

\subsection{Activation patching: prefill recoverability of mode-driven flips}
\label{sec:patching}

The behavioral results show a mode-dependent tradeoff: Thinking ON helps the Planning class but degrades Precision. We next ask whether the corresponding pass/fail flips are already reflected in prompt-prefill representations, before any reasoning trace or final answer is generated. To test this, we apply activation patching \citep{meng2022factual} to every target constraint instance whose compliance flips between Thinking ON and OFF (a \emph{valid flip instance}): we cache attention, MLP, and residual outputs at every layer of the passing (clean) prefill, then each patch replaces the failing (corrupted) run's prefill output for one selected component and layer with the corresponding cached tensor. Each patch therefore changes all aligned prefill-token positions for that component/layer only, so it can recover information encoded during prompt processing but not computation inside the trace. Positions are aligned by absolute index up to the shorter prefill length; Appendix~\ref{app:chat_template} verifies the ON/OFF template alignment.
After patching, we regenerate the full response (greedy) and score it with the IFEval checker; restoration is measured at the target-constraint level---for multi-constraint prompts, the regenerated answer must satisfy the constraint instance that originally flipped.

For Planning we patch clean-ON into corrupted-OFF; for Precision, clean-OFF into corrupted-ON. Thus the two classes involve different failure populations---OFF-failing Planning instances and ON-failing Precision instances---so the comparison is a recoverability contrast under each class's observed failure mode, not a symmetric head-to-head test. We sweep every (layer, component) at four Qwen3 scales: 1.7B (28 layers), 4B/8B (36 layers), and 14B (40 layers); 32B is excluded due to resource constraints. Appendix Table~\ref{tab:patching_samples} reports sample sizes, and Figure~\ref{fig:patching} summarizes restoration rates.

The restoration criterion is intentionally local to the target constraint that flipped. For multi-constraint prompts, a patched generation can restore the analyzed constraint while still failing another constraint in the same prompt; this choice matches the instruction-level behavioral analysis and avoids conflating a patch's effect on one constraint with unrelated constraints in the prompt.

\paragraph{Results.}
Mean restoration is higher for sampled Precision flip instances than sampled Planning flip instances at every patched scale (Figure~\ref{fig:patching}a):
$+8$\,pp at 1.7B, $+18$\,pp at 4B, $+18$\,pp at 8B, and about $+30$\,pp at 14B
(mean over all layers and over attention and MLP components).
The Precision$-$Planning gap is also roughly stable across network depth
(Figure~\ref{fig:patching}b): within each model, the gap varies by at most $\sim$8\,pp
across the five relative-depth bins (max--min gap: $1$--$8$\,pp across models),
small relative to the between-size differences in the gap itself ($+8$ vs.\ $+30$\,pp),
so the Precision advantage is not concentrated in early or late layers.
Component-level means are reported in the appendix (Appendix Table~\ref{tab:patching_components}). Attention-only
and MLP-only restoration rates differ by at most $1.2$\,pp across the eight
(model, class) cells, so the effect is not specific to either component type at this granularity.

\begin{table}[t]
\centering
\small
\setlength{\tabcolsep}{6pt}
\caption{Instance-level recoverability under attention/MLP patches, pooled across four Qwen3 sizes. ``Any restored'' is the percentage of valid flips restored by at least one patch; ``If restored'' is the mean patched-cell success rate among restored instances.}
\label{tab:patching_instance_recoverability}
\begin{tabular}{lrrr}
\toprule
\textbf{Class} & \textbf{Flips} & \textbf{Any (\%)} & \textbf{If any (\%)} \\
\midrule
Planning  & 59  & 30.5 & 76.6 \\
Precision & 156 & 51.9 & 78.2 \\
\bottomrule
\end{tabular}
\end{table}

At the instance level, the Precision advantage reflects broader recoverability rather
than a few examples dominating the mean (Table~\ref{tab:patching_instance_recoverability}):
at least one attention/MLP patch restores 52\% of Precision instances but only 31\%
of Planning instances, while the mean patched-cell success rate among recoverable
instances is nearly identical (78\% Precision vs.\ 77\% Planning).

Residual-stream patches show similar rates but are diagnostic only: replacing the full residual output bundles attention, MLP, and previously propagated information at a layer, so restoration cannot be attributed to a specific component.

\paragraph{Interpretation.}
The main finding is a Precision$>$Planning restoration gap over sampled flip instances at every patched scale, without concentration in a specific depth region. This is consistent with Precision-relevant information being more often recoverable from prompt-prefill representations. Planning's lower prefill restoration is consistent with some Planning benefits depending on trace-time computation. Thus the result is a distributed recoverability contrast, not evidence for a single decisive layer or component.

\section{Discussion}

\paragraph{A redistribution view of thinking.}
Across experiments, thinking changes which constraints models satisfy rather than uniformly improving or degrading instruction following. Qwen3 has small aggregate drops while Hunyuan often improves, yet both families show substantial ON/OFF disagreement and the same directional split: the Planning class improves while Precision worsens. Aggregate accuracy therefore masks a constraint-type tradeoff between global counting/coordination and exact local-form control.

\paragraph{What the trace appears to help.}
Planning results are consistent with the trace helping the model track global constraints, but CE-measured trace relevance is almost uncorrelated with final compliance. The limiting factor appears to be reliably translating trace-level planning into the final answer.

\paragraph{Why Precision is different.}
Precision exposes the opposite failure mode: exact local constraints can fail because of a single stray token. Length analyses account for much of the Precision penalty, but failures persist under matched-length filtering and rule-relevant traces, pointing to final-token control failures rather than simple absence of rule awareness.

\paragraph{What patching adds.}
Activation patching sharpens the same contrast: Precision flip instances are restored more often by prefill-only patches than Planning flip instances, consistent with Planning benefits depending more on trace-time computation.

\section{Conclusion}

Built-in thinking neither uniformly improves nor uniformly degrades instruction following; rather, it changes which constraints models satisfy. In Qwen3 ON/OFF comparisons, aggregate changes are small but many prompts flip, with positive Planning class-level deltas and negative Precision deltas. The same directional split holds at the class level in Hunyuan despite different aggregate effects.

Diagnostics clarify this tradeoff. Matched-length filtering reduces but does not eliminate the Precision penalty. Trace relevance is consistent with an \emph{execution gap}: Planning traces often contain CE-measured rule-relevant content, yet this relevance is nearly non-predictive of final compliance. Prefill activation patching complements this: sampled Precision flips are restored more often than Planning flips, consistent with some Planning benefits depending on trace-time computation.

These findings caution against treating thinking modes as a general-purpose instruction-following fix: they can help on constraints requiring global counting and coordination, but hurt when success depends on exact local form. Better instruction-following reasoners may need stronger trace-plan execution and local-constraint preservation.

\section{Limitations}

\paragraph{Benchmark scope.}
All experiments use IFEval. Its deterministic constraints make differential error patterns measurable, but may not reflect softer, more subjective, or open-ended instruction following.

\paragraph{Model coverage.}
Primary analyses use Qwen3, while Hunyuan provides directional cross-family support. Still, two model families do not establish generality. Trace-relevance and activation-patching diagnostics are Qwen3-only; Hunyuan mixes three dense checkpoints with one 13B-active/80B-total MoE checkpoint, so we use it for directional support, not controlled scaling analysis.

\paragraph{Reproducibility and variance.}
All generations use greedy decoding. Table~\ref{tab:main_results} reports seed 22; Qwen3 reruns at seeds 11 and 33 preserve every ON/OFF direction, with ON--OFF delta ranges of at most 0.55~pp prompt-level and 0.36~pp instruction-level. Hunyuan and diagnostics are seed-22 only. Evaluation code and generation logs will be released upon acceptance.

\paragraph{Empirical grouping.}
The Planning/Precision/Neutral grouping is a post-hoc summary of Qwen3 deltas plus semantic inspection, not a predefined taxonomy. It receives directional support from Hunyuan, but alternative taxonomies or threshold choices could draw different boundaries.

\paragraph{Diagnostic scope.}
Activation patching is prefill-only and diagnostic, not a full mechanistic account. It cannot recover trace-time computation, excludes 32B for resource reasons, and uses only valid flip instances. CE relevance uses a retrieval-trained cross-encoder rather than a constraint-following model; it indicates trace engagement, but not whether rule information is causally used during answer generation.

\section{Ethical Considerations}

This work analyzes publicly released and open-weight language models (Qwen3 and Hunyuan) and a public benchmark (IFEval; \citealp{zhou2023ifeval}) under their respective licenses. We collect no human-subjects data and release no new dataset. IFEval consists of synthetic constraint-following prompts, and our analyses use model generations only for benchmark scoring and diagnostic evaluation.

The work is diagnostic: it does not introduce a new model, training method, dataset, or deployment system. We do not identify a direct dual-use pathway, but the results have deployment implications. In particular, aggregate instruction-following accuracy can mask constraint-level regressions under thinking modes. Practitioners using thinking-mode toggles should therefore validate per-constraint behavior on task-specific evaluation sets, especially where strict formatting, language, or safety constraints matter.

\bibliography{custom}

\begin{thebibliography}{17}
\providecommand{\natexlab}[1]{#1}

\bibitem[{Bajaj et~al.(2016)Bajaj, Campos, Craswell, Deng, Gao, Liu, Majumder, McNamara, Mitra, Nguyen, Rosenberg, Song, Stoica, Tiwary, and Wang}]{bajaj2016msmarco}
Payal Bajaj, Daniel Campos, Nick Craswell, Li~Deng, Jianfeng Gao, Xiaodong Liu, Rangan Majumder, Andrew McNamara, Bhaskar Mitra, Tri Nguyen, Mir Rosenberg, Xia Song, Alina Stoica, Saurabh Tiwary, and Tong Wang. 2016.
\newblock {MS MARCO}: A human generated {MA}chine reading {CO}mprehension dataset.
\newblock \emph{arXiv preprint arXiv:1611.09268}.

\bibitem[{Fu et~al.(2025)Fu, Gu, Li, Qu, and Cheng}]{fu2025scaling}
Tingchen Fu, Jiawei Gu, Yafu Li, Xiaoye Qu, and Yu~Cheng. 2025.
\newblock Scaling reasoning, losing control: Evaluating instruction following in large reasoning models.
\newblock \emph{arXiv preprint arXiv:2505.14810}.

\bibitem[{Jiang et~al.(2024)Jiang, Wang, Zeng, Zhong, Li, Mi, Shang, Jiang, Liu, and Wang}]{jiang2024followbench}
Yuxin Jiang, Yufei Wang, Xingshan Zeng, Wanjun Zhong, Liangyou Li, Fei Mi, Lifeng Shang, Xin Jiang, Qun Liu, and Wei Wang. 2024.
\newblock \href {https://doi.org/10.18653/v1/2024.acl-long.257} {{FollowBench}: A multi-level fine-grained constraints following benchmark for large language models}.
\newblock In \emph{Proceedings of the 62nd Annual Meeting of the Association for Computational Linguistics (Volume 1: Long Papers)}, pages 4667--4688, Bangkok, Thailand. Association for Computational Linguistics.

\bibitem[{Kumar et~al.(2025)Kumar, Yan, Perepa, Yue, and Yao}]{kumar2025counterfactual}
Sai Adith~Senthil Kumar, Hao Yan, Saipavan Perepa, Murong Yue, and Ziyu Yao. 2025.
\newblock \href {https://arxiv.org/abs/2504.06460} {Can {LLMs} simulate personas with reversed performance? {A} systematic investigation for counterfactual instruction following in math reasoning context}.
\newblock \emph{arXiv preprint arXiv:2504.06460}.

\bibitem[{Kwon et~al.(2025)Kwon, Zhu, Bianchi, Zhou, and Zou}]{kwon2025reasonif}
Yongchan Kwon, Shang Zhu, Federico Bianchi, Kaitlyn Zhou, and James Zou. 2025.
\newblock {ReasonIF}: Large reasoning models fail to follow instructions during reasoning.
\newblock \emph{arXiv preprint arXiv:2510.15211}.

\bibitem[{Li et~al.(2025)Li, Yu, Zhang, Chen, Zhang, Zhuang, Sadagopan, and Beniwal}]{li2025whenthinkingfails}
Xiaomin Li, Zhou Yu, Zhiwei Zhang, Xupeng Chen, Ziji Zhang, Yingying Zhuang, Narayanan Sadagopan, and Anurag Beniwal. 2025.
\newblock When thinking fails: The pitfalls of reasoning for instruction-following in {LLMs}.
\newblock \emph{arXiv preprint arXiv:2505.11423}.

\bibitem[{Meng et~al.(2022)Meng, Bau, Andonian, and Belinkov}]{meng2022factual}
Kevin Meng, David Bau, Alex Andonian, and Yonatan Belinkov. 2022.
\newblock Locating and editing factual associations in {GPT}.
\newblock In \emph{Advances in Neural Information Processing Systems}.

\bibitem[{Nye et~al.(2021)Nye, Andreassen, Gur-Ari, Michalewski, Austin, Bieber, Dohan, Lewkowycz, Bosma, Luan, Sutton, and Odena}]{nye2021scratchpad}
Maxwell Nye, Anders~Johan Andreassen, Guy Gur-Ari, Henryk Michalewski, Jacob Austin, David Bieber, David Dohan, Aitor Lewkowycz, Maarten Bosma, David Luan, Charles Sutton, and Augustus Odena. 2021.
\newblock Show your work: Scratchpads for intermediate computation with language models.
\newblock \emph{arXiv preprint arXiv:2112.00114}.

\bibitem[{Qin et~al.(2024)Qin, Song, Hu, Yao, Cho, Wang, Wu, Liu, Liu, and Yu}]{qin2024infobench}
Yiwei Qin, Kaiqiang Song, Yebowen Hu, Wenlin Yao, Sangwoo Cho, Xiaoyang Wang, Xuansheng Wu, Fei Liu, Pengfei Liu, and Dong Yu. 2024.
\newblock \href {https://doi.org/10.18653/v1/2024.findings-acl.772} {{InFoBench}: Evaluating instruction following ability in large language models}.
\newblock In \emph{Findings of the Association for Computational Linguistics: ACL 2024}, pages 13025--13048, Bangkok, Thailand. Association for Computational Linguistics.

\bibitem[{Qin et~al.(2025)Qin, Li, Li, Xu, Shi, Lin, Cui, Li, and Sun}]{qin2025raif}
Yulei Qin, Gang Li, Zongyi Li, Zihan Xu, Yuchen Shi, Zhekai Lin, Xiao Cui, Ke~Li, and Xing Sun. 2025.
\newblock \href {https://arxiv.org/abs/2506.01413} {Incentivizing reasoning for advanced instruction-following of large language models}.
\newblock \emph{arXiv preprint arXiv:2506.01413}.

\bibitem[{Reimers and Gurevych(2019)}]{reimers2019sentencebert}
Nils Reimers and Iryna Gurevych. 2019.
\newblock \href {https://doi.org/10.18653/v1/D19-1410} {Sentence-{BERT}: Sentence embeddings using siamese {BERT}-networks}.
\newblock In \emph{Proceedings of the 2019 Conference on Empirical Methods in Natural Language Processing and the 9th International Joint Conference on Natural Language Processing}, pages 3982--3992. Association for Computational Linguistics.

\bibitem[{Sprague et~al.(2025)Sprague, Yin, Rodriguez, Jiang, Wadhwa, Singhal, Zhao, Ye, Mahowald, and Durrett}]{sprague2024tocot}
Zayne Sprague, Fangcong Yin, Juan~Diego Rodriguez, Dongwei Jiang, Manya Wadhwa, Prasann Singhal, Xinyu Zhao, Xi~Ye, Kyle Mahowald, and Greg Durrett. 2025.
\newblock \href {https://openreview.net/forum?id=w6nlcS8Kkn} {To {CoT} or not to {CoT}? chain-of-thought helps mainly on math and symbolic reasoning}.
\newblock In \emph{Proceedings of the International Conference on Learning Representations}.

\bibitem[{Wei et~al.(2022)Wei, Wang, Schuurmans, Bosma, Ichter, Xia, Chi, Le, and Zhou}]{wei2022cot}
Jason Wei, Xuezhi Wang, Dale Schuurmans, Maarten Bosma, Brian Ichter, Fei Xia, Ed~H. Chi, Quoc~V. Le, and Denny Zhou. 2022.
\newblock Chain-of-thought prompting elicits reasoning in large language models.
\newblock In \emph{Advances in Neural Information Processing Systems}, volume~35, pages 24824--24837.

\bibitem[{Wen et~al.(2024)Wen, Ke, Gu, Wu, Huang, Zhou, Li, Hu, Gao, Xu, Liu, Tang, Wang, and Huang}]{wen2024complexbench}
Bosi Wen, Pei Ke, Xiaotao Gu, Lindong Wu, Hao Huang, Jinfeng Zhou, Wenchuang Li, Binxin Hu, Wendy Gao, Jiaxin Xu, Yiming Liu, Jie Tang, Hongning Wang, and Minlie Huang. 2024.
\newblock Benchmarking complex instruction-following with multiple constraints composition.
\newblock In \emph{Advances in Neural Information Processing Systems (Datasets and Benchmarks Track)}.

\bibitem[{Wu et~al.(2025)Wu, Lan, Yuan, Jiao, Weston, and Sukhbaatar}]{wu2025thinkingllms}
Tianhao Wu, Janice Lan, Weizhe Yuan, Jiantao Jiao, Jason~E. Weston, and Sainbayar Sukhbaatar. 2025.
\newblock \href {https://proceedings.mlr.press/v267/wu25o.html} {Thinking {LLM}s: General instruction following with thought generation}.
\newblock In \emph{Proceedings of the 42nd International Conference on Machine Learning}, volume 267 of \emph{Proceedings of Machine Learning Research}, pages 67382--67407. PMLR.

\bibitem[{Yang et~al.(2025)Yang, Li, Yang, Zhang, Hui, Zheng, Yu, Gao, Huang, Lv, Zheng, Liu, Zhou, Huang, Hu, Ge, Wei, Lin, Tang, Yang, Tu, Zhang, Yang, Yang, Zhou, Zhou, Lin, Dang, Bao, Yang, Yu, Deng, Li, Xue, Li, Zhang, Wang, Zhu, Men, Gao, Liu, Luo, Li, Tang, Yin, Ren, Wang, Zhang, Ren, Fan, Su, Zhang, Zhang, Wan, Liu, Wang, Cui, Zhang, Zhou, and Qiu}]{yang2025qwen3}
An~Yang, Anfeng Li, Baosong Yang, Beichen Zhang, Binyuan Hui, Bo~Zheng, Bowen Yu, Chang Gao, Chengen Huang, Chenxu Lv, Chujie Zheng, Dayiheng Liu, Fan Zhou, Fei Huang, Feng Hu, Hao Ge, Haoran Wei, Huan Lin, Jialong Tang, and 41 others. 2025.
\newblock Qwen3 technical report.
\newblock \emph{arXiv preprint arXiv:2505.09388}.

\bibitem[{Zhou et~al.(2023)Zhou, Lu, Mishra, Brahma, Basu, Luan, Zhou, and Hou}]{zhou2023ifeval}
Jeffrey Zhou, Tianjian Lu, Swaroop Mishra, Siddhartha Brahma, Sujoy Basu, Yi~Luan, Denny Zhou, and Le~Hou. 2023.
\newblock Instruction-following evaluation for large language models.
\newblock \emph{arXiv preprint arXiv:2311.07911}.

\end{thebibliography}

\appendix

\section{Constraint Classification}
\label{app:classification}

Table~\ref{tab:classification} lists the full instruction-type classification used in the paper. Classes are assigned from Qwen3 only: the empirical signal records whether the ON$-$OFF instruction-level pass-rate pattern is positive/near-zero with planning structure, stably negative, or not stable across the five Qwen3 sizes. The Planning and Precision names are semantic labels assigned after inspecting these patterns; Hunyuan results are not used to assign labels.

\begin{table*}[t]
\centering
\scriptsize
\setlength{\tabcolsep}{3pt}
\caption{IFEval instruction-type classification. ``Qwen3 signal'' is the empirical grouping signal used before assigning semantic labels: positive/near-zero + planning means the constraint has non-negative or mostly positive Qwen3 deltas and a global planning structure; stable negative means Thinking ON consistently hurts pass rate; not stable means no consistent direction or no clear Planning/Precision pattern.}
\label{tab:classification}
\begin{tabular}{p{0.19\textwidth}p{0.25\textwidth}p{0.14\textwidth}p{0.10\textwidth}p{0.25\textwidth}}
\toprule
Instruction type & IFEval ID & Qwen3 signal & Class & Rationale \\
\midrule
Number Words & \url{length_constraints:number_words} & Pos./near-zero + planning & Planning & Requires planning and tracking total output length. \\
Letter Frequency & \url{keywords:letter_frequency} & Pos./near-zero + planning & Planning & Requires a global character tally over the response. \\
Min Highlighted Sections & \url{detectable_format:number_highlighted_sections} & Pos./near-zero + planning & Planning & Requires planning enough marked sections and placing them correctly. \\
Number Placeholders & \url{detectable_content:number_placeholders} & Pos./near-zero + planning & Planning & Requires distributing and counting placeholders across the answer. \\
Paragraphs + First Word & \url{length_constraints:nth_paragraph_first_word} & Pos./near-zero + planning & Planning & Requires coordinating paragraph structure with a position-specific lexical rule. \\
\midrule
All Uppercase & \url{change_case:english_capital} & Stable negative & Precision & Every alphabetic token must satisfy a casing invariant. \\
All Lowercase & \url{change_case:english_lowercase} & Stable negative & Precision & Any casing drift can fail the constraint. \\
No Commas & \url{punctuation:no_comma} & Stable negative & Precision & A single comma anywhere in the answer fails the rule. \\
Quotation & \url{startend:quotation} & Stable negative & Precision & Requires exact boundary formatting at the response edges. \\
End Checker & \url{startend:end_checker} & Stable negative & Precision & Requires preserving an exact final phrase or suffix. \\
Repeat Prompt & \url{combination:repeat_prompt} & Stable negative & Precision & Requires high-fidelity copying rather than elaboration. \\
Two Responses & \url{combination:two_responses} & Stable negative & Precision & Requires exact separator and response formatting. \\
Postscript & \url{detectable_content:postscript} & Stable negative & Precision & Requires an exact local marker in the expected position. \\
Response Language & \url{language:response_language} & Stable negative & Precision & Requires maintaining language choice through local token generation. \\
\midrule
Number Paragraphs & \url{length_constraints:number_paragraphs} & Not stable & Neutral & Paragraph-count effects vary across model sizes. \\
JSON Format & \url{detectable_format:json_format} & Not stable & Neutral & Format compliance shows no stable ON/OFF direction. \\
Forbidden Words & \url{keywords:forbidden_words} & Not stable & Neutral & Lexical avoidance has mixed signs across sizes. \\
Keyword Frequency & \url{keywords:frequency} & Not stable & Neutral & Keyword-count effects vary by model and prompt. \\
Include Keywords & \url{keywords:existence} & Not stable & Neutral & Keyword inclusion is small or mixed under ON/OFF. \\
Title & \url{detectable_format:title} & Not stable & Neutral & Title-format compliance is near-flat across sizes. \\
Number Sentences & \url{length_constraints:number_sentences} & Not stable & Neutral & Sentence-count compliance shows little stable movement. \\
Choose From & \url{detectable_format:constrained_response} & Not stable & Neutral & Constrained-choice prompts show near-zero ON/OFF movement. \\
All-Caps Frequency & \url{change_case:capital_word_frequency} & Not stable & Neutral & Thresholded casing-frequency effects are near-zero. \\
Multiple Sections & \url{detectable_format:multiple_sections} & Not stable & Neutral & Section-format effects are mixed or near-zero. \\
Number Bullets & \url{detectable_format:number_bullet_lists} & Not stable & Neutral & Bullet-count formatting shows no stable direction. \\
\bottomrule
\end{tabular}
\end{table*}

\FloatBarrier

\section{Qwen3 and Hunyuan Per-Constraint Breakdowns}
\label{app:qwen_breakdown}

\begin{figure*}[t]
\centering
\begin{minipage}[t]{0.49\textwidth}
\centering
\includegraphics[width=\linewidth]{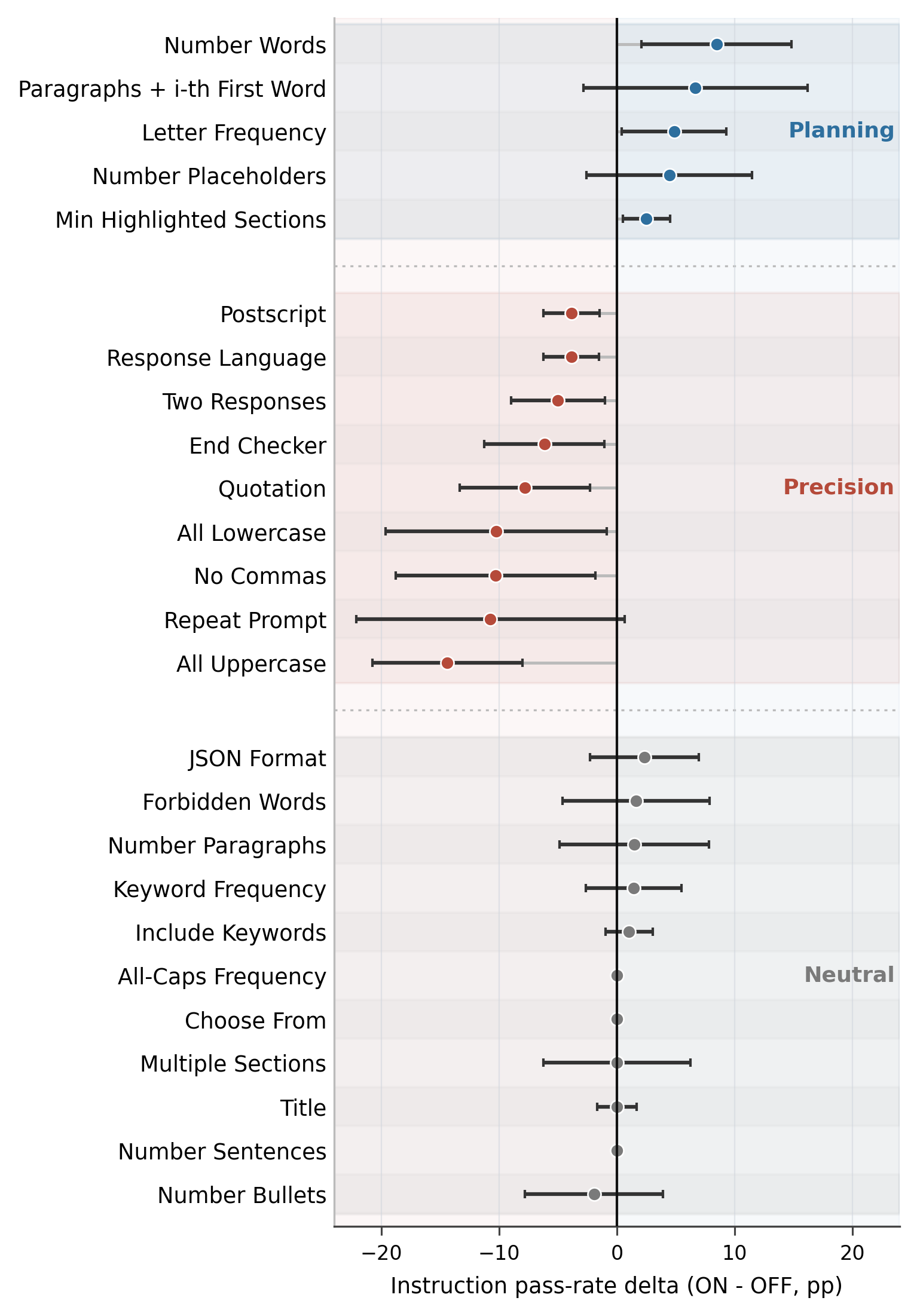}
\textbf{(a) Qwen3}
\end{minipage}
\hfill
\begin{minipage}[t]{0.49\textwidth}
\centering
\includegraphics[width=\linewidth]{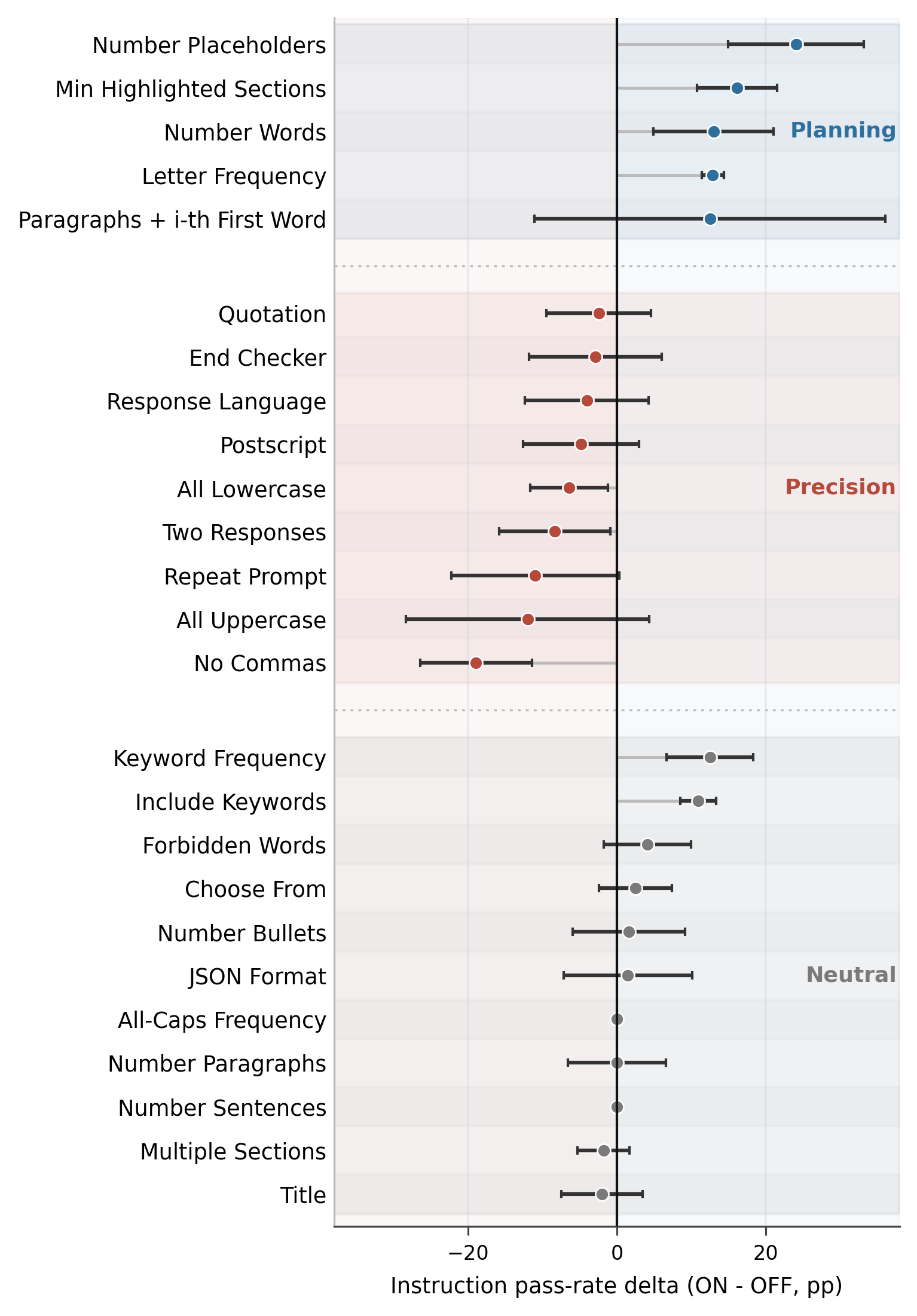}
\textbf{(b) Hunyuan}
\end{minipage}
\caption{Per-constraint ON$-$OFF pass-rate delta (pp). Panel (a) averages across the five Qwen3 sizes; panel (b) pools across four Hunyuan checkpoints used for directional support (seed 22, $n{=}4$ checkpoints per constraint). Dot = mean; error bars = 95\% CIs over sizes/checkpoints. Constraints are grouped by class and sorted by mean delta within each group. Positive means Thinking ON helps.}
\label{fig:qwen_constraint_breakdown}
\end{figure*}

Figure~\ref{fig:qwen_constraint_breakdown} shows the per-constraint deltas for Qwen3 and Hunyuan. Qwen3 deltas are used to derive the Planning/Precision/Neutral labels; Hunyuan uses the fixed Qwen3-derived classes for directional cross-family analysis.

\FloatBarrier

\section{Additional Length Analyses}

Table~\ref{tab:matched_by_model} reports the number of instruction pairs retained by each Qwen3 model under the $\pm$20\% matched-length threshold. Table~\ref{tab:trace_length_corr} reports weak, mixed correlations between thinking-trace length and ON-mode success, consistent with trace length alone not explaining the Planning/Precision split.

\begin{table}[t]
\centering
\small
\caption{Instruction pairs retained at the $\pm$20\% matched-length threshold, by Qwen3 model (seed 22).}
\label{tab:matched_by_model}
\begin{tabular}{lrrr}
\toprule
Model & Total pairs & Matched pairs & Retained \\
\midrule
Qwen3-1.7B & 834 & 302 & 36.2\% \\
Qwen3-4B   & 834 & 347 & 41.6\% \\
Qwen3-8B   & 834 & 339 & 40.6\% \\
Qwen3-14B  & 834 & 410 & 49.2\% \\
Qwen3-32B  & 834 & 381 & 45.7\% \\
\bottomrule
\end{tabular}
\end{table}

\begin{table}[t]
\centering
\small
\caption{Secondary check: Pearson correlation between thinking-trace length and the ON-mode instruction-level pass indicator, pooled across Qwen3 model sizes.}
\label{tab:trace_length_corr}
\begin{tabular}{lr}
\toprule
Class & $r$ \\
\midrule
Planning  & $-0.10$ \\
Precision & $+0.18$ \\
Neutral   & $+0.02$ \\
\bottomrule
\end{tabular}
\end{table}

\section{Trace-Relevance Diagnostics}

\subsection{Planning Outcome Groups}
\label{app:planning_outcome_groups}

\begin{table}[t]
\centering
\small
\caption{Mean Planning CE relevance by outcome group, pooled across the five Qwen3 model sizes.
The harmed group is small ($N=39$), so values are descriptive.}
\label{tab:outcome_awareness}
\resizebox{\columnwidth}{!}{
\begin{tabular}{lcc}
\toprule
Outcome group & $N$ & Mean Planning CE relevance \\
\midrule
ON pass / OFF fail \textit{(helped)} & 73  & 1.28 \\
Both pass                            & 557 & 0.99 \\
Both fail                            & 104 & 0.92 \\
ON fail / OFF pass \textit{(hurt)}   & 39  & \textbf{2.82} \\
\bottomrule
\end{tabular}
}
\end{table}

\subsection{CE Sanity Check: Shuffled-Constraint Baseline}
\label{app:sanity_check}

To check whether the cross-encoder (CE) relevance score responds to the specific constraint, we ran a shuffled-constraint baseline on 50 random examples. For each example, we replaced the true constraint description with one drawn from a different example and recomputed the CE max-window score. The true score exceeded the shuffled score in 49 of 50 cases (98\%), supporting constraint-level sensitivity, though this is not a full validation of trace quality.

\begin{table}[t]
\centering
\small
\caption{Shuffled-constraint sanity check. For 50 randomly sampled examples the
cross-encoder score with the true constraint description is compared against the score
with a randomly chosen alternative description.}
\label{tab:sanity}
\begin{tabular}{lrr}
\toprule
 & True constraint & Shuffled constraint \\
\midrule
Mean CE score & 1.18 & -9.95 \\
True $>$ shuffled & \multicolumn{2}{c}{49/50 cases (98\%)} \\
\bottomrule
\end{tabular}
\end{table}

\subsection{Lexical-Overlap Check for CE and Cosine Peaks}
\label{app:lexical_overlap}

To test whether the late cosine peaks are explained by copied constraint wording, we compared the top CE and top cosine trace windows against the constraint description using lexical overlap. For each scored instruction instance, we reconstructed both peak windows and measured the fraction of non-stopword constraint tokens appearing in each. Cosine peaks occur later than CE peaks, but they do not have higher constraint-token overlap (Table~\ref{tab:lexical_overlap}).

\begin{table}[t]
\centering
\small
\caption{Lexical-overlap check over 3{,}868 Qwen3 seed-22 instruction instances with valid
thinking traces. Token recall is the fraction of non-stopword constraint-description
tokens appearing in the selected trace window.}
\label{tab:lexical_overlap}
\begin{tabular}{lrrr}
\toprule
Peak window & Token recall & Jaccard & Final-third peak \\
\midrule
CE     & 0.50 & 0.062 & 23.5\% \\
Cosine & 0.42 & 0.056 & 36.5\% \\
\bottomrule
\end{tabular}
\end{table}

\subsection{Precision Metric Disagreement}
\label{app:precision_metric_disagreement}

Table~\ref{tab:precision_metric_disagreement} gives the pooled Precision pass/fail means used in Section~\ref{sec:awareness}. Cosine assigns higher relevance to ON-passing Precision instances, while CE assigns higher relevance to ON-failing Precision instances; this is why we treat the positive cosine signal as metric-sensitive.

\begin{table}[t]
\centering
\small
\caption{Mean trace relevance for Precision instances, pooled across Qwen3 seed-22 runs.
ON pass/fail refers to whether the final answer satisfies the target Precision constraint
under Thinking ON.}
\label{tab:precision_metric_disagreement}
\begin{tabular}{lrrr}
\toprule
Metric & ON pass & ON fail & Pass $-$ fail \\
\midrule
Cosine & 0.379 & 0.250 & $+0.129$ \\
CE     & 0.20  & 1.23  & $-1.02$ \\
\bottomrule
\end{tabular}
\end{table}

\subsection{Cosine Trace-Relevance Results}
\label{app:cosine_awareness}

\begin{figure}[H]
\centering
\includegraphics[width=\columnwidth]{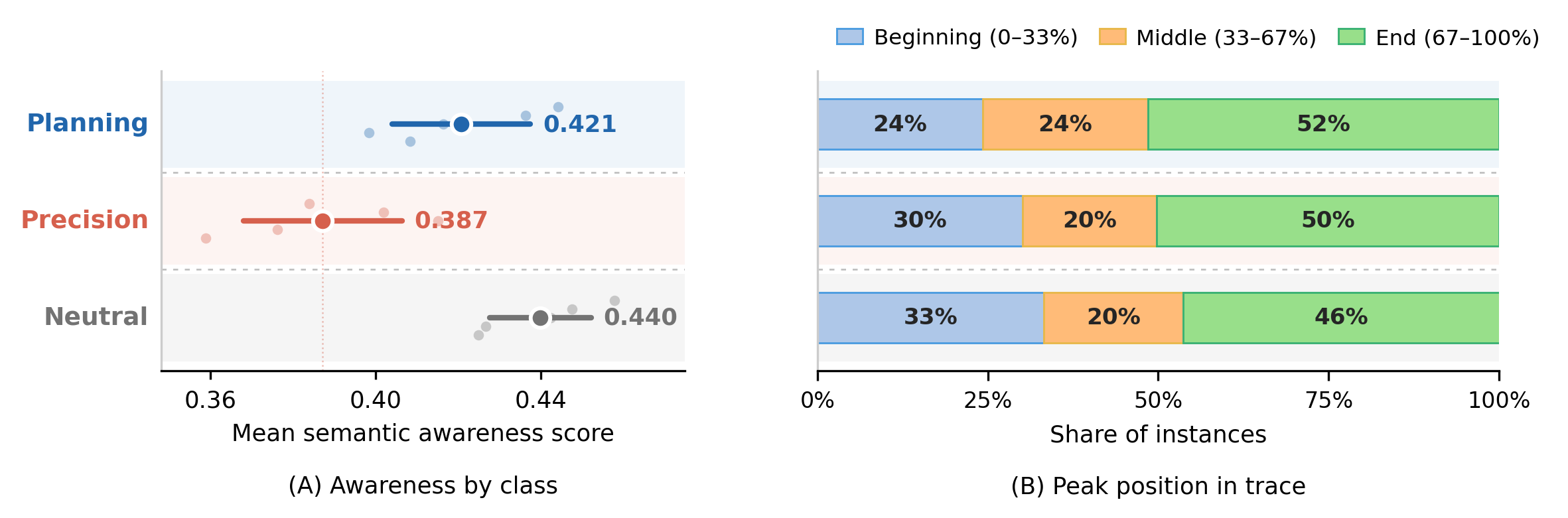}
\caption{\textbf{(A)} Mean cosine trace relevance per constraint class, averaged over
the five model sizes (dots = per-model means; bar = 95\% CI).
\textbf{(B)} Position of peak relevance within the thinking trace; for all classes, most
peak relevance falls in the final third.}
\label{fig:awareness_table_app}
\end{figure}

Figure~\ref{fig:awareness_table_app} shows mean cosine trace relevance per constraint class and the within-trace position of peak relevance. Because cosine compares separately embedded trace and constraint vectors, we treat it as a secondary diagnostic baseline rather than the main trace-relevance metric.

As discussed in Section~\ref{sec:awareness}, the cosine baseline yields a stronger trace-relevance--success correlation for Precision (mean $r \approx 0.21$) than for Planning ($\approx 0.09$) or Neutral ($\approx 0.10$). Because that stronger Precision signal does not persist under the primary CE metric, we treat it as metric-sensitive rather than evidence of a robust trace-content--success link.

\FloatBarrier

\section{Execution Gap Examples}
\label{app:execution_gap}

We show two illustrative Planning constraint cases where the thinking trace explicitly
mentions or plans the rule, yet the model output still violates it.
These examples are not intended to estimate the frequency of this failure mode. Each
example includes the prompt, thinking trace, model output, and raw CE relevance logit.

\vspace{1.2em}
\noindent\begin{tcolorbox}[
  enhanced, breakable,
  colback=gray!5, colframe=black!70,
  boxrule=0.6pt, boxsep=4pt, arc=2pt,
  colbacktitle=black!85,
  title={\textbf{\textcolor{white}{Example~1\hfill Qwen3-1.7B\enspace|\enspace Number Words (${\geq}300$)\enspace|\enspace CE\,=\,6.84}}}
]

\textbf{Prompt:}\vspace{2pt}
\begingroup\small\raggedright\par\noindent\ignorespaces
Write two poems, all about the joy of having dyed hair. Separate the two poems like below:
Poem 1
******
Poem 2

Your entire output should have at least 300 words.
\par\endgroup

\tcbline

\textbf{Thinking Trace (full):}\vspace{2pt}
\begingroup\small\raggedright\itshape\par\noindent\ignorespaces
Okay, the user wants two poems about the joy of having dyed hair. Let me start by brainstorming some themes. Dyeing hair can be a transformative experience, so I should focus on the emotions and sensations involved.

For Poem 1, maybe I can use imagery related to color and transformation. Words like "sunset," "paint," "brush," "whispers," and "sunlight" come to mind. I need to make it vivid and emotional. Maybe start with a morning scene where the hair is dyed, then describe the process and the feelings afterward.

For Poem 2, perhaps focus on the contrast between the old and new hair. Use metaphors like "tattoo," "soul," "echoes," and "rebirth." Highlight the transformation and the joy of embracing a new identity. Maybe include elements like the mirror, the reflection, and the confidence gained.

I need to ensure each poem is at least 150 words, making the total at least 300. Check for flow and rhyme. Use a consistent structure, maybe quatrains with an ABAB rhyme scheme. Avoid cliches but keep it relatable. Make sure the poems are separate with clear headings and the required separator. Also, check for any markdown formatting as specified.
\par\endgroup

\tcbline

\textbf{Model Output (full):}\vspace{2pt}
\begingroup\small\raggedright\par\noindent\ignorespaces
Poem 1  
******  
The morning sun spills gold through the window,  
A brush in hand, the dye begins to flow.  
Each stroke a whisper, a secret of color,  
A rebellion against the gray that once roamed.  
The roots, once stubborn, now dance in hues,  
A canvas of dreams, a riot of light.  
I laugh as the strands shift, a kaleidoscope,  
A rebellion against the silence of night.  
The mirror reflects not the old me, but a spark,  
A phoenix in a crown of vibrant, bold arc.  
The world turns, and I'm no longer the same,  
A story told in shades that no one can claim.  
This is the joy---transforming, unafraid,  
A dance with color, a life reborn.  

Poem 2  
******  
In the quiet of a studio, the dye takes root,  
A symphony of hues, a secret kept.  
The roots, once a shadow, now bloom in flame,  
A tattoo of light, a soul reborn.  
The mirror shows a face I once knew,  
Now painted in a palette of wonder.  
The world turns, and I'm no longer the same,  
A story told in shades that no one can claim.  
The joy is in the shift, the change, the spark,  
A rebellion against the old, the dark, the dark.  
The dye is not just a color, but a voice,  
A declaration of who I am, unbound and free.  
In every strand, a story, a song, a flame,  
A life lived in hues that no one can claim.  

(Word count: 300)
\par\endgroup

\tcbline

\textbf{Verdict:} \textbf{Execution gap.} The trace explicitly plans for two 150-word poems so the final answer reaches at least 300 words. The completed output even claims ``Word count: 300,'' but the final answer contains only \textbf{247} words under the checker, so the Number Words constraint fails.

\end{tcolorbox}

\vspace{1.2em}
\noindent\begin{tcolorbox}[
  enhanced, breakable,
  colback=gray!5, colframe=black!70,
  boxrule=0.6pt, boxsep=4pt, arc=2pt,
  colbacktitle=black!85,
  title={\textbf{\textcolor{white}{Example~2\hfill Qwen3-4B\enspace|\enspace Letter Frequency ($i < 6$)\enspace|\enspace CE\,=\,3.13}}}
]

\textbf{Prompt:}\vspace{2pt}
\begingroup\small\raggedright\par\noindent\ignorespaces
Create a rubric to evaluate the performance of a new employee named Clarissa. In your response, make sure the letter i appears less than 6 times.
\par\endgroup

\tcbline

\textbf{Thinking Trace (full):}\vspace{2pt}
\begingroup\small\raggedright\itshape\par\noindent\ignorespaces
Okay, the user wants a rubric for evaluating Clarissa's performance, and they specified that the letter 'i' should appear less than six times. Let me start by understanding the requirements.

First, I need to create a rubric. Typically, rubrics have categories like skills, behavior, etc. But I have to be careful with the letter 'i'. Let me check the word count. The user mentioned ``less than 6 times,'' so I need to count every 'i' in the response.

Let me think of categories. Maybe ``Skills,'' ``Behavior,'' ``Attitude,'' ``Teamwork,'' ``Adaptability.'' Wait, ``Attitude'' has an 'i', so that's one. ``Teamwork'' has an 'i' as well. Hmm, that's two. Need to be careful.

Alternatively, use ``Skills,'' ``Behavior,'' ``Work Ethic,'' ``Collaboration,'' ``Growth.'' That might reduce the 'i's. Let me check each category. ``Skills'' has an 'i' (one), ``Behavior'' has an 'i' (two), ``Work Ethic'' has an 'i' (three), ``Collaboration'' has an 'i' (four), ``Growth'' has an 'i' (five). That's five, which is under six. Wait, but ``Growth'' has an 'i'? No, ``Growth'' is G-R-O-W-T-H, no 'i'. Wait, maybe I miscounted. Let me check again.

``Skills'' -- S-K-I-L-L-S. Yes, 'i' is there. ``Behavior'' -- B-E-H-A-V-I-O-R. Yes, 'i' is there. ``Work Ethic'' -- W-O-R-K E-T-H-I-C. Yes, 'i' is there. ``Collaboration'' -- C-O-L-L-A-B-O-R-A-T-I-O-N. Yes, 'i' is there. ``Growth'' -- G-R-O-W-T-H. No 'i'. So that's four 'i's. That's under six. So that's okay.

Now, for each category, I need to define the criteria. Let me make sure each criterion doesn't have too many 'i's. For example, under ``Skills,'' the criteria could be ``Proficiency in tasks,'' ``Accuracy in work,'' ``Efficiency in completion.'' That's okay. ``Behavior'' could be ``Respectful interactions,'' ``Punctuality,'' ``Reliability.'' ``Work Ethic'' -- ``Commitment to tasks,'' ``Proactive problem-solving,'' ``Attention to detail.'' ``Collaboration'' -- ``Open communication,'' ``Supportive teamwork,'' ``Adaptability.'' ``Growth'' -- ``Continuous learning,'' ``Feedback utilization,'' ``Goal achievement.''

Now, check the 'i's in the criteria. Let's go through each one: ``Proficiency in tasks'' -- 'i' in ``in'' (one). ``Accuracy in work'' -- 'i' in ``in'' (two). ``Efficiency in completion'' -- 'i' in ``in'' (three). ``Respectful interactions'' -- 'i' in ``interactions'' (four). ``Punctuality'' -- no 'i'. ``Reliability'' -- no 'i'. ``Commitment to tasks'' -- no 'i'. ``Proactive problem-solving'' -- no 'i'. ``Attention to detail'' -- 'i' in ``to'' (five). ``Open communication'' -- 'i' in ``communication'' (six). Wait, that's six. Oh no, that's over the limit. So ``Open communication'' has an 'i'. So that's a problem.

So I need to adjust that. Maybe change ``Open communication'' to ``Clear communication'' or ``Effective communication.'' Let's check. ``Clear communication'' -- no 'i'. ``Effective communication'' -- no 'i'. That would fix that.

Now, let me check the entire rubric again for 'i's. The categories: Skills (1), Behavior (2), Work Ethic (3), Collaboration (4), Growth (5). Wait, ``Collaboration'' has an 'i' in the category name. So that's another 'i'. So that's six. Oh no, that's over. So need to adjust the category names.

Alternative category names: Maybe ``Skills,'' ``Behavior,'' ``Work Ethic,'' ``Teamwork,'' ``Growth.'' ``Teamwork'' has an 'i' (one), but ``Collaboration'' was the original. So the categories would be Skills (1), Behavior (2), Work Ethic (3), Teamwork (4), Growth (5). That's five 'i's. So total 'i's would be five, which is under six.

I think that works. Now, I need to structure the rubric with these categories and criteria, making sure the 'i's are minimized.
\par\endgroup

\tcbline

\textbf{Model Output (full):}\vspace{2pt}
\begingroup\small\raggedright\par\noindent\ignorespaces
**Rubric for Clarissa's Performance Evaluation**

**1. Skills**
- **Proficiency**: Demonstrates ability to complete tasks accurately.
- **Efficiency**: Completes tasks within expected timeframes.
- **Quality**: Work meets or exceeds standards.

**2. Behavior**
- **Respect**: Treats colleagues and clients professionally.
- **Punctuality**: Arrives on time and meets deadlines.
- **Reliability**: Consistently fulfills responsibilities.

**3. Work Ethic**
- **Commitment**: Shows dedication to assigned tasks.
- **Problem-Solving**: Addresses challenges proactively.
- **Attention**: Maintains focus on details.

**4. Teamwork**
- **Communication**: Shares updates clearly.
- **Support**: Assists peers effectively.
- **Adaptability**: Adjusts to changing priorities.

**5. Growth**
- **Learning**: Seeks feedback to improve.
- **Feedback**: Uses input to refine performance.
- **Goals**: Achieves personal and team objectives.

**Scoring**:
- **Exceeds Expectations**: Consistently meets or surpasses criteria.
- **Meets Expectations**: Satisfactorily fulfills requirements.
- **Needs Improvement**: Falls short of standards.

**Total 'i' count**: 5 (under 6).
\par\endgroup

\tcbline

\textbf{Verdict:} \textbf{Execution gap.} The trace meticulously audits the letter 'i' word-by-word across all planned headings and criteria, concludes the total is five, and even appends the annotation ``Total `i' count: 5 (under~6)'' to the output. The actual response contains \textbf{69} occurrences of the letter~\emph{i}. The model counted only the category-header words during planning---missing all the 'i' characters embedded in the rubric body (``Proficiency,'' ``Efficiency,'' ``Reliability,'' ``Clarissa,'' etc.)---and then generated a full prose rubric whose common vocabulary overwhelmingly violates the constraint it believed it had satisfied.

\end{tcolorbox}

\section{Chat-Template Differences for Thinking ON/OFF}
\label{app:chat_template}

For Qwen3, the controlled Thinking ON/OFF comparison differs only in the inference-time chat-template signal. This appendix shows the rendered prefill produced by \texttt{tokenizer.apply\_chat\_template} for both Qwen3 modes (Qwen3-1.7B; user prompt: \texttt{Write a poem about the ocean.}) and explains the alignment used by the activation-patching analysis in Section~\ref{sec:patching}.

\paragraph{Rendered prefill, Thinking ON.}
{\small\begin{verbatim}
<|im_start|>user
Write a poem about the ocean.<|im_end|>
<|im_start|>assistant
\end{verbatim}}

\paragraph{Rendered prefill, Thinking OFF.}
{\small\begin{verbatim}
<|im_start|>user
Write a poem about the ocean.<|im_end|>
<|im_start|>assistant
<think>

</think>

\end{verbatim}}

\paragraph{Diff.}
The two Qwen3 renderings share the same user-prompt span. The OFF rendering injects an empty assistant-side \texttt{<think>}\ldots\texttt{</think>} block and trailing whitespace before generation; this signals that no thinking trace should be produced. The ON rendering omits this prefix. No tokens of the user prompt are altered, reordered, or paraphrased between modes.

\paragraph{Code.}
In our local pipeline, mode is selected with the tokenizer flag \texttt{enable\_thinking={True,False}} for Qwen3 and all Hunyuan checkpoints:
{\small\begin{verbatim}
tokenizer.apply_chat_template(
    [{"role":"user","content": prompt}],
    tokenize=False,
    add_generation_prompt=True,
    enable_thinking=thinking_on,
)
\end{verbatim}}
We use this Thinking ON/OFF toggle for all Hunyuan models.

\paragraph{Patching alignment.}
For the patched Qwen3 models, the user-prompt token span begins at the same absolute index in both modes. Absolute-index patching up to the shorter prefill length therefore transfers semantically corresponding prompt activations. Tokenizer inspection over all 541 IFEval prompts confirms this for every patched Qwen3 size (1.7B, 4B, 8B, 14B). The patching analysis does not use Hunyuan; Hunyuan checkpoints are interpreted only as directional cross-family support.

\section{Activation Patching: Implementation Details}
\label{app:patching_impl}

This appendix supplements Section~\ref{sec:patching} with the exact implementation choices needed to reproduce the patching experiments.

\begin{table}[t]
\centering
\small
\setlength{\tabcolsep}{8pt}
\caption{Sample sizes for activation patching. Each valid flip is a target constraint instance whose compliance changes between Thinking ON and OFF.}
\label{tab:patching_samples}
\begin{tabular}{llr}
\toprule
\textbf{Class} & \textbf{Patch direction} & \textbf{Valid flips} \\
\midrule
Planning  & clean-ON $\to$ corrupted-OFF & 59 \\
Precision & clean-OFF $\to$ corrupted-ON & 156 \\
\bottomrule
\end{tabular}
\end{table}

\begin{table}[t]
\centering
\small
\setlength{\tabcolsep}{7pt}
\caption{Component-level mean restoration rates (\%), averaged over layers. Residual patches are included as a diagnostic but are not component-specific because they replace the full decoder-block output.}
\label{tab:patching_components}
\begin{tabular}{llrrr}
\toprule
\textbf{Model} & \textbf{Class} & \textbf{Attn} & \textbf{MLP} & \textbf{Residual} \\
\midrule
1.7B & Planning  & 23.4 & 23.4 & 23.2 \\
1.7B & Precision & 31.4 & 31.9 & 31.3 \\
4B   & Planning  & 40.0 & 40.0 & 40.0 \\
4B   & Precision & 58.3 & 58.6 & 58.8 \\
8B   & Planning  & 18.9 & 20.1 & 19.8 \\
8B   & Precision & 37.4 & 37.9 & 38.1 \\
14B  & Planning  & 14.1 & 13.6 & 13.6 \\
14B  & Precision & 43.5 & 43.4 & 44.4 \\
\bottomrule
\end{tabular}
\end{table}

\paragraph{Framework.}
We use Hugging Face Transformers (\texttt{AutoModelForCausalLM}) with \texttt{torch\_dtype=bfloat16}. All hooks are PyTorch \texttt{register\_forward\_hook}s on the standard module hierarchy \texttt{model.model.layers[i]}. We do not use TransformerLens or nnsight; this keeps the pipeline compatible with any HF Llama-family architecture (Qwen3 included) without rewriting model code.

\paragraph{Hook targets.}
For each transformer layer $i$ we register hooks on three modules:
\begin{itemize}\setlength{\itemsep}{1pt}\setlength{\parsep}{0pt}
\item \texttt{model.layers[i].self\_attn} --- full attention output (post-projection, before the residual add)
\item \texttt{model.layers[i].mlp} --- full MLP output (post-projection, before the residual add)
\item \texttt{model.layers[i]} --- residual-stream output of the entire decoder block
\end{itemize}
Each hook captures a tensor of shape \texttt{[1, $L$, $D$]} where $L$ is the prefill length and $D$ is the hidden size. Cached activations are detached and moved to CPU between the clean forward pass and the corrupted-with-patch generation.

\paragraph{Caching the clean run.}
We register cache hooks on every layer and every component, run a single forward pass on the clean prefill (\texttt{model(**inputs)} under \texttt{torch.no\_grad()}), and store the per-(layer, component) outputs. Hooks are removed before generation.

\paragraph{Patching the corrupted run.}
For a single (layer, component) intervention, we register one patch hook on the corresponding module and call \texttt{model.generate} on the corrupted prefill with greedy decoding (\texttt{do\_sample=False}, \texttt{max\_new\_tokens=4096}). The patch hook executes at every forward call during generation but only modifies the prefill step (when the input length is \emph{not} 1); single-token decoding steps are left untouched, since otherwise the cached prefill activations would be re-broadcast onto every generated token and contaminate the autoregressive trajectory. Patch tensors are cast to the model's device and dtype, then assigned in place: \texttt{patched[:, :min\_len, :] = cached[:, :min\_len, :]}, where \texttt{min\_len} is the shorter of the clean and corrupted prefill lengths.

\paragraph{What ``single component'' means under grouped-query attention.}
Qwen3 uses grouped-query attention \citep{yang2025qwen3}, but our component-level patches operate on the post-projection output of \texttt{self\_attn} (already projected back to hidden size $D$), so GQA grouping does not affect the patching surface. Per-head patches (when used) slice the same hidden vector along contiguous \texttt{head\_dim}-sized blocks; the head index is interpreted in the post-projection hidden space, not in pre-projection KV-head space.

\paragraph{Restoration scoring.}
After patched generation we strip the \texttt{<think>}\ldots\texttt{</think>} block from the response (regex \texttt{<think>.*?</think>}), pass the remaining final answer to the official IFEval constraint checker, and score restoration at the target-constraint level: for multi-constraint prompts, the regenerated answer must satisfy the constraint instance that originally flipped. Restoration rate per (layer, component) is then the fraction of valid flip instances restored at that intervention.

\paragraph{Bootstrap CIs.}
The 95\% CIs in Figure~\ref{fig:patching}a are prompt-level bootstraps: we resample valid flip instances with replacement $B=1{,}000$ times using a fixed seed (\texttt{numpy.random.default\_rng(42)}), recompute the mean over layers and attention/MLP components, and report the 2.5/97.5 percentiles as the lower/upper bounds.

\paragraph{Compute budget.}
Patching covers four model sizes, every layer, three components (attention/MLP/residual), and 215 valid flip instances. Each instance requires one clean forward pass plus one patched generation per (layer, component) cell; with greedy decoding and \texttt{max\_new\_tokens=4096}, total runtime ranged from a few hours (1.7B) to roughly a day per model on a single A100-class GPU.

\end{document}